# The "Right" Discourse on Migration: Analysing Migration-Related Tweets in Right and Far-Right Political Movements

Nishan Chatterjee[1,2,5], Veronika Bajt[3], Ana Zwitter Vitez[4], Senja Pollak[5]

[1]University of La Rochelle, La Rochelle, France, [2]Jožef Stefan International Postgraduate School, Ljubljana, Slovenia - nishan_chatterjee@outlook.com
[3]The Peace Institute, Ljubljana, Slovenia - Veronika.Bajt@mirovni-institut.si
[4]University of Ljubljana, Ljubljana, Slovenia - ana.zwitter@guest.arnes.si
[5]Jožef Stefan Institute, Ljubljana, Slovenia - senja.pollak@ijs.si

## Abstract

The rise of right-wing populism in Europe has brought to the forefront the significance of analysing social media discourse to understand the dissemination of extremist ideologies and their impact on political outcomes. Twitter, as a platform for interaction and mobilisation, provides a unique window into the everyday communication of far-right supporters. In this paper, we propose a methodology that uses state-of-the-art natural language processing techniques with sociological insights to analyse the MIGR-TWIT corpus of far-right tweets in English and French. We aim to uncover patterns of discourse surrounding migration, hate speech, and persuasion techniques employed by right and far-right actors. By integrating linguistic, sociological, and computational approaches, we seek to offer cross-disciplinary insights into societal dynamics and contribute to a better understanding of contemporary challenges posed by right-wing extremism on social media platforms.

**Keywords:** social media, far-right discourse, natural language processing, topic modelling, hate speech

## 1. Introduction

Research shows the advance of right-wing populism in Europe (Brubaker, 2017), thus analysing right and far-right tweets is important for understanding the dissemination of extremist ideologies, as well as how mobilisation of supporters occurs towards achieving electoral success (Kaiser et al., 2019). A platform like Twitter is used for "dissemination, interaction, mobilisation and building the personality of influencer/leaders and as a strategic tool for the selection of issues and for propaganda and fake news" (Pérez, 2020). Examining online discourse in contrast to traditional methods such as interviewing means non-reactive data that is "given" and as such not influenced by researchers (Muis et al., 2020). Twitter enables unobtrusive insight into real-life everyday discussions among far-right supporters, while at the same time providing political leaders the opportunity to circumvent traditional news channels, making their voices heard without interference. Underlining the significance of studying far-right social media activity, highlighting its role in shaping public opinion, and influencing political outcomes, social research is interested in the potential impact on social cohesion and political discourse. A study by Marwick et al., (2017) found that far-right groups have developed techniques of "attention hacking" to increase the visibility of their ideas through the strategic use of social media, memes, and bots. By targeting journalists and influencers to help spread content, the importance of analysing online extremist communities to identify patterns of radicalization and to develop effective countermeasures is underscored. Examining right and far-right tweets is hence crucial for addressing contemporary societal challenges. Also, analyses of far-right online activities can contribute to the identification of radicalization patterns and help inform strategies to counter extremism (Marwick et al., 2017).

Empirical research (Mos, 2019) shows, however, that to avoid getting expelled from the platform, both far-right political parties and right-wing extremist movements have adapted their tone to Twitter's policy.

Qualitative analyses play a crucial role in understanding societal dynamics and can be used to examine discourses by individuals or institutions (Van Raalte et al., 2021, Ahmed et al., 2021). Yet qualitative analyses are time-consuming, constrained to small datasets, and can rely on pre-formulated hypotheses. Hence, with the availability of large data collections and computational methods, many studies nowadays rely on the use of natural language processing techniques (NLP) on large Twitter datasets for sentiment analysis (Tumasjan et al., 2010, Jaki et al., 2019), hate speech detection (Talat et al., 2016, Espinosa et al., 2019), network analysis (Conover et al., 2012, Lima et al., 2018), or extremist content identification (e.g., Hartung et al., 2017), but rarely integrate social sciences and linguistics perspectives for analysing the results. However, some studies use computational methods and large corpora to better understand far-right discourse. Vidgen et al. (2021) use machine learning and latent Markov modelling on data of the British National Party followers on Twitter, showing far-right actors are often purveyors of hate speech, using social media to spread divisive and prejudiced messages that inflict harm on targeted victims, create a sense of fear in communities, stir intergroup tensions and conflict in far-right ideologies expressed on social media platforms which can serve as indicators of political sentiments and public opinion. Conover et al. (2011) combine network clustering algorithms and manually annotated data, emphasizing the significance of analysing political communication on social media for gauging ideological divides. Froio and Ganesh (2017) propose a methodological framework combining quantitative methods (social network analysis and logistic regression) and qualitative discourse analysis to analyse the transnationalisation of far-right discourse on Twitter.

In our study, we use the MIGR-TWIT Corpus by Battaglia et al. (2022), a bilingual diachronic dataset capturing tweets from 2011 to 2022 in British and French political contexts. Pietrandrea and Battaglia (2022) have used this data to scrutinise the linguistic evolution of migration discourse between the right and far-right-wing French politicians. Through corpus methods and critical discourse analysis, their work uncovers the semantic shifts and manipulative strategies employed by the right in associating migration with diverse and unrelated topics across a decade. To the best of our knowledge, this is the only publicly available tweet-corpus on migrations across 10 years. We propose a methodology using a set of state-of-the-art NLP techniques—neural topic modelling, cross-lingual hate speech categorisation and persuasion-techniques classifier—to the corpus of (far-)right tweets. We believe that NLP in combination with linguistic and sociological interpretations enables a better understanding of right-wing discourses.

## 2. Data

We used the right and far-right political actors' tweets from the MIGR-TWIT v2 corpus in English and French (Battaglia et al. 2022; Sangwan, 2023, Pietrandrea and Battaglia, 2022). The UK sub-corpus in our study covers the period 2012-2022 and consists of 6,472 tweets posted by 12 political parties or figures. Most tweets were posted by the UK Independence Party, a Eurosceptic right-wing populist political party that peaked in the mid-2010s when it was the largest party representing the UK in the European Parliament. Following was Nigel Farage, the UKIP leader (2006-2009, 2010-2016) and leader of the Brexit Party (renamed Reform UK in 2021). Tweets by the UK Home Office come in third, a department of the British Government, responsible for immigration, security, and law and order.

The French sub-corpus covers 2011-2022, comprising 16 user accounts (11,761 tweets). The National Rally (formerly known as the National Front) led the pack, advocating for anti-immigration policies and the protection of French identity. Marine Le Pen's account, a prominent figure within the party, closely follows, as does Nicolas Bay, known for his involvement in far-right movements. Many other politicians and organisations appear in the dataset, including Éric Zemmour, recognized for his controversial views on immigration.

## 3. Methodology

Our methodology consists of three state-of-the-art NLP techniques.

### *3.1 Topic Modelling*

Topic modelling enables the extraction of the underlying topic clusters, which would be manually impractical and time-consuming. We use the novel BERTopic (Grootendorst, 2022) leveraging pre-trained embeddings and multilingual datasets, surpassing older methods like NMF and LDA (Egger et al., 2022). BERTopic's flexibility in supporting various topic modelling variations and its effectiveness in handling multilingual data also make it our preferred choice. We utilized the multilingual-e5-large Sentence Transformer model (Wang et al., 2022) and the following BERTopic settings: UMAP for dimensionality reduction (with 15 neighbours and 10 components for the UK and 15 neighbours and 5 components for the FR tweets); HDBSCAN for clustering (min. cluster size of 25 for the UK and 45 for the FR sub-corpus); and Class-based TF-IDF for cluster-specific keyword extraction as well as KeyBERT topics. While TF-IDF provides global keywords to uniquely identify each cluster, KeyBERT topics provide an alternate interpretation of the topics representing the document clusters through Maximal Marginal Relevance. The differing settings in the UMAP and HDBSCAN are due to the differing sizes of the two sub-corpora.

To automatically generate descriptive topic labels for each cluster (labels in English), we prompted LLama2 (Touvron et al., 2023) with a custom prompt along with the TF-IDF cluster keywords and the 3 most-representative tweets from each cluster. Since LLama2 contains a few safety guardrails, some of the topic labels were initially deemed too hateful to be generated, which was resolved using structured output guidelines and nudging the model with "Certainly I'll be happy to do so. The label for this topic is: ". BERTopic also provides methods to visualize the evolution of topics and their representative keywords over time.

### *3.3 Toxicity in Tweets*

To identify harmful tweets, we use an off-the-shelf multilingual toxicity classifier (Hanu et al., 2020) with a classification accuracy of 92.11 on the Jigsaw Multilingual Toxic Comment Classification task (Kivlichan et al., 2020). It additionally labels toxic tweets with the following categories: obscenity, threat, insult, identity attacks, and sexual explicitness.

### *3.4 Persuasion in Tweets*

Understanding persuasion strategies in tweets is crucial for analysing the dynamic interplay between social media content from political movements, user engagement, and public attention. Schwemmer (2021) highlights that the nature of the message conveyed significantly influences audience response. We fine-tune an mBART-50 model (Tang et al., 2020) on the textual subpart of the SemEval 2024 Task 4 data on online disinformation campaigns (Dimitrov et al. 2024) for identifying 20 hierarchically organised persuasion techniques. The model we used (Chatterjee, 2024) reaches a 0.62 hierarchical F1 score on the shared tasks

English test set, with a potential decrease of up to 10% for corpora of similar sizes for languages not encountered during training.

## 4. Results

### 4.1 Topic Analysis

In the UK sub-corpus, among the 39 topics detected (see Fig. 2), the most frequent topic talks about "Nigel Farage's views on immigration and border control" (Freq. 442) The tweets are composed of comments on immigration figures or warnings about an influx of immigrants to the UK (Ex. 1.1 and 1.2, Fig. 1), relating this supposed threat to Britain's EU membership and offering a solution in leaving the EU (Ex. 1.3, Fig. 1). The element of border control is seen as something impossible to achieve (Ex. 1.1, Fig. 1) while also as the moment when control over the UK territory could be regained - but only on exiting the EU (Ex. 1.3, Fig. 1).

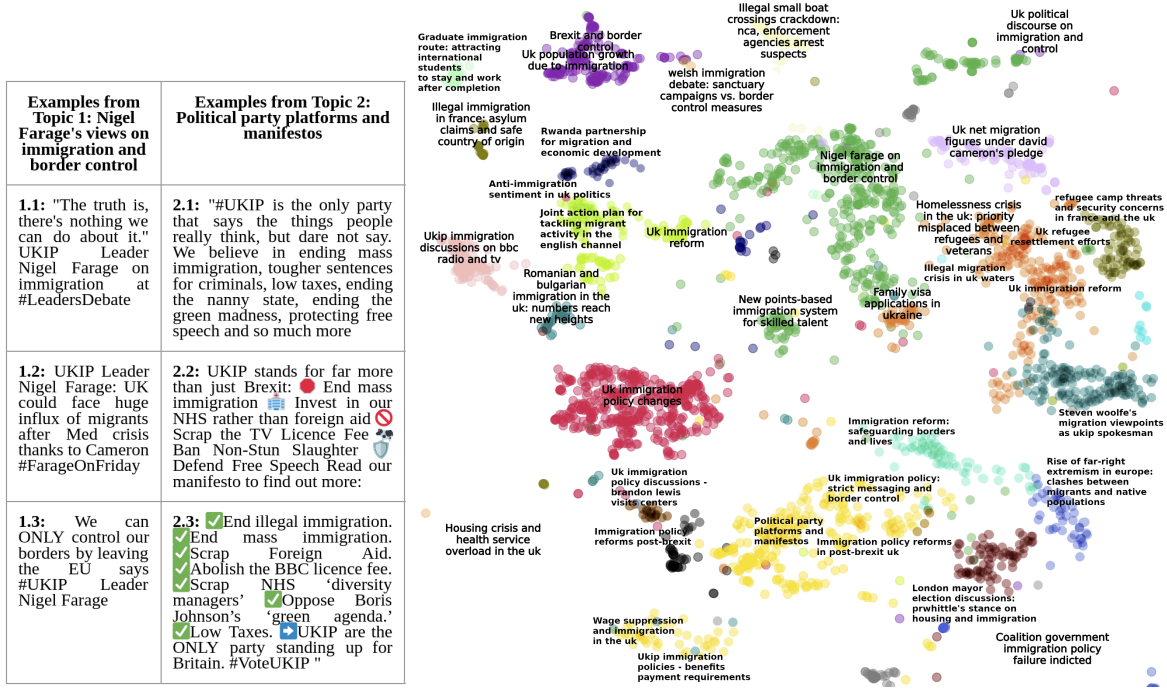

| Examples from Topic 1: Nigel Farage's views on immigration and border control | Examples from Topic 2: Political party platforms and manifestos |
|---|---|
| **1.1:** "The truth is, there's nothing we can do about it." UKIP Leader Nigel Farage on immigration at #LeadersDebate | **2.1:** "#UKIP is the only party that says the things people really think, but dare not say. We believe in ending mass immigration, tougher sentences for criminals, low taxes, ending the nanny state, ending the green madness, protecting free speech and so much more |
| **1.2:** UKIP Leader Nigel Farage: UK could face huge influx of migrants after Med crisis thanks to Cameron #FarageOnFriday | **2.2:** UKIP stands for far more than just Brexit: 🔴 End mass immigration 🗳️ Invest in our NHS rather than foreign aid 🚫 Scrap the TV Licence Fee 🔪 Ban Non-Stun Slaughter 🛡️ Defend Free Speech Read our manifesto to find out more: |
| **1.3:** We can ONLY control our borders by leaving the EU says #UKIP Leader Nigel Farage | **2.3:** ✅End illegal immigration. ✅End mass immigration. ✅Scrap Foreign Aid. ✅Abolish the BBC licence fee. ✅Scrap NHS 'diversity managers' ✅Oppose Boris Johnson's 'green agenda.' ✅Low Taxes. ➡️UKIP are the ONLY party standing up for Britain. #VoteUKIP " |

Fig 1: Examples from UK Tweets    Fig 2: Llama2 Topic Labels for tweet-clusters in UK Tweets[1]

The second most frequent topic is "Political party platforms and manifestos" (Freq. 440). The keywords, such as: 'labour', 'ukip', and 'vote' refer to UKIP's main political opposition coming from the Labour Party and the importance of gaining their respective supporters' votes. It is interesting to find also 'brexit' and 'immigration' in BERTopic keywords, which allows a great summation of the main preoccupation of the political debate with leaving the EU and immigration. We can see the typical right-wing populist emphasis on daring to speak the truth, protecting people's right to free speech, and clamping down on immigration and crime. Being against ecology (Ex. 2.1 and 2.3, Fig. 1), and public broadcaster's licence fee is also emphasized (Ex. 2.2 and 2.3, Fig. 1). The latter is typically the case in populist political discourse, where politicians often adopt the victim's perspective, supposedly being on the receiving end of mainstream's media "fake news" (e.g. Donald Trump).

---

[1] The dynamic visualization of topic clusters, topic evolution, detected persuasion and toxicity, along with the code, can be found at github.com/nishan-chatterjee/the-right-discourse.git.

The third most frequent topic, "Rise of far-right extremism in Europe: clashes between migrants and native populations" is also interesting. The TF-IDF keywords, such as ['liberal', 'elite', 'liberals', 'europe', 'diversity', 'greatreplacement', 'soros', 'enrichment', 'war', 'refugees'] refer to common populist fearmongering tactics about George Soros, liberals, and diversity leading to the so-called Great Replacement, a white nationalist far-right conspiracy theory in which white European populations are being demographically and culturally replaced by non-white migrants/refugees, especially from Muslim countries. BERTopic keywords refer to similar concepts, e.g. 'refugees', 'whitegenocide', 'invasion', 'islamist', 'greatreplacement', 'racism', adding also 'rapists' that goes hand in hand with the racialized view of (supposedly Muslim) migrant men.

The diachronic view of themes in the UK corpus (see Fig. 3) reveals a densification of peaks around the time of the so-called refugee crisis that engulfed Europe in 2015. The top three topics "Nigel Farage on immigration and border control", "Political party platforms and manifestos", "Rise of far-right extremism in Europe: clashes between migrants and native populations" all have peaks in 2015. The latter, though, has the highest peak in 2017 which can be seen as related to a widely mediatized attack in Germany that involved an Afghan asylum seeker. Also peaking in 2015 is the fourth most frequent topic "Brexit and border control", as well as several smaller topics about immigration, policy reforms, and political discourse. Moreover, the theme "ISIS terrorist infiltration via migrant surge threatens European national security" was non-existent before 2015, when it suddenly erupted with keywords such as ['isis', 'fighters', 'jihadist', 'islamic', 'terrorists', 'tide', 'use', 'flood', 'europe', 'extremists'].

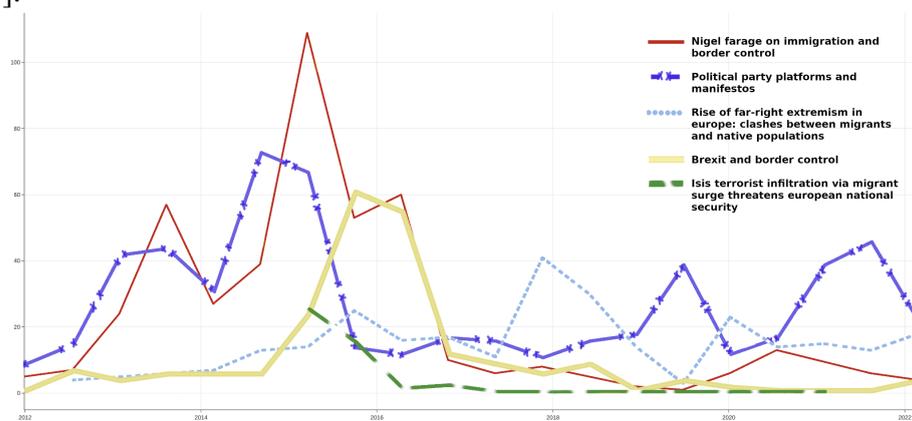

Fig 3: Evolution of Topics over time for the UK Tweets

In the French sub-corpus, from the detected topics (see Fig. 5), the most frequently identified topics tend to advocate against the EU and French national migration policies which promote the idea of a multicultural society and "demographic submersion" (topics "French political discourse on immigration and EU policies", "Immigration-crime link in urban areas", "French migration policy debate"). The authors of tweets assert that the EU and Macron's government represent a threat to their identity because the current direction may lead to the disappearance of European nations (Ex. 1, Fig. 4). The sentiments of threat and fear are also evident in the topic "Immigration crime in French urban areas" (Ex. 2, Fig. 4).

The topic of the threat to European identity is closely tied to the opportunity to gain popularity among local voters. This is evident in topics such as "French political discourse on immigration: referendum proposal" and "Bardella's migrant crisis discourse", exploited by Marine Le Pen, the leader of Rassemblement National (National Rally) at the time of corpus compilation, and Jordan Bardella, spokesperson of the same political party during the corpus compilation. In their posts, they often propose a referendum to halt the historical French right of Droit du sol (birthright citizenship), which grants French citizenship to every baby born on French territory (Ex. 3, Fig. 4). Other keywords also reveal an interesting parallel between French far-right politicians in Italy (Matteo Salvini) and Hungary (Viktor Orbán).

While in migration-related literature the topics related to threat and fear are frequently discussed, the topics of "Controversial migrant debate in France: pushback efforts, economic motives, and Human trafficking" and "Turkish aggression threaten European borders" are interesting as they contain formulations promoting empathy, directed either towards migrants (Ex. 4, Fig. 4) or towards local authorities further intensifying animosity towards migrants (Ex. 5, Fig. 4).

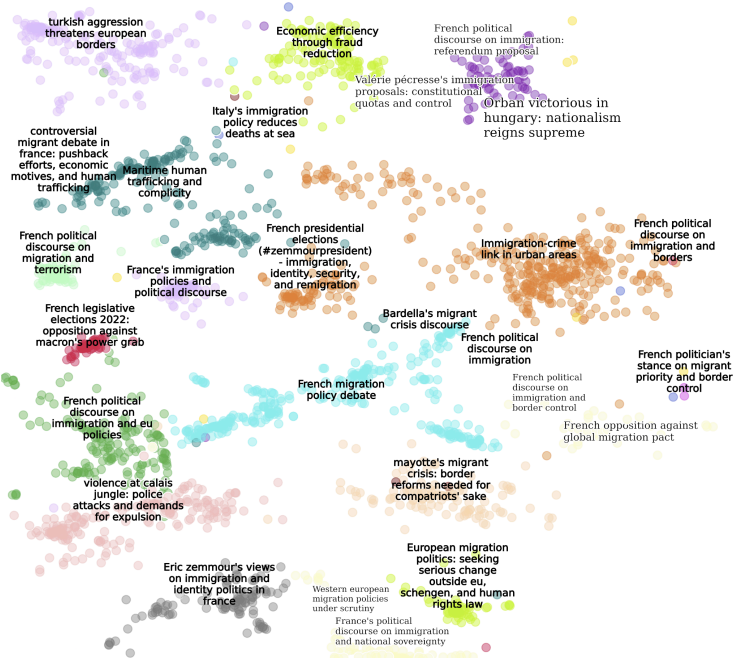

Fig 4: Examples from French Tweets    Fig 5: Llama2 Topic Labels for tweet-clusters in French Tweets

The diachronic overview of themes in Fig. 6 reveals three significant peaks: 2017 (with a smaller peak in 2015) for the topic of "French Political discourse on immigration and EU policies", 2018 for "Bardella's migrant crisis discourse", and 2021 for "French political discourse on immigration: referendum proposal". The topic related to EU migration policies emerged simultaneously with the "refugee crisis" in 2015, marked by a large influx of people from the Middle East and North Africa. Security concerns intensified with a series of terrorist attacks in France during this period, including the Charlie Hebdo attack, the attacks in Paris, and the Nice truck attack. The French presidential election in 2017 was also strongly marked by debates on immigration, explaining why a smaller topic, "France's political discourse on immigration and national sovereignty", also peaked in 2017. In 2018, the global refugee and humanitarian crisis persisted, but a new wave of protests, the Yellow Vest movement, emerged in France, focusing on various issues, including economic inequality and dissatisfaction with government policies. This represented an opportunity for far-right politicians, such as Jordan Bardella, to solidify their position by addressing issues related to immigration, national identity, and general discontent. It is not surprising that other, smaller topics, such as "Maritime human trafficking and complicity" or "Macron's immigration policies: national vs. european frontiers", reached their peak in 2018. The last peak, related to the topic "French political discourse on immigration: referendum proposal", seems to be continually growing from 2020 onwards (since the Covid-19 pandemic). It refers to M. Le Pen's project to discontinue the historical French right of "Droit du sol," which grants French citizenship to every baby born on French territory. This topic is accompanied by a less prominent topic, "French migration policy debate", showing a similar increasing trend.

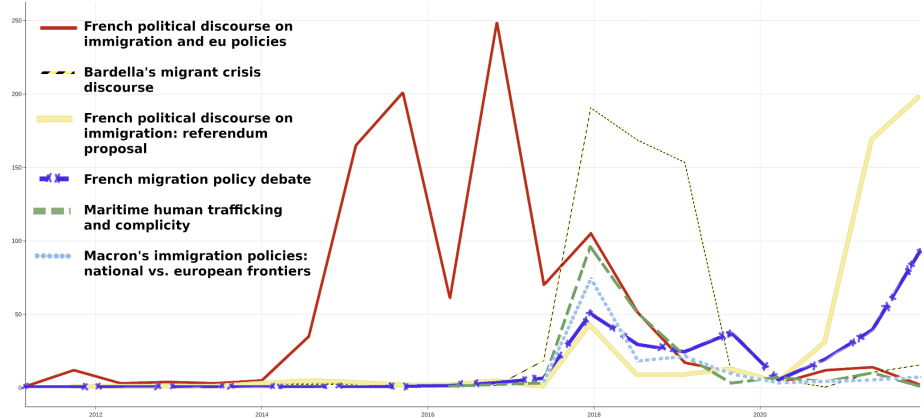

Fig 6: Evolution of Topics over Time for the French Tweets

## *4.3 Toxicity in Tweets*

Using the multilingual toxic-comment classifier, we extract interesting examples for qualitative analysis (in total, 163 and 107 for UK and FR, respectively). In the first category *of Obscenity,* we mostly find vulgar words ("bullshit européen" – European bullshit, "pédé" - faggot), but also drug dealing (Ex. 1, Fig. 7).

| |
|---|
| **1. Obscenity:** La génération de mes parents a fait l'effort d'apprendre le français, de respecter les lois, de travailler. Aujourd'hui, une certaine immigration vient pour dealer du shit, agresser les femmes, cracher sur la France : ces gens-là, nous devons les renvoyer chez eux. [English: My parents' generation made the effort to learn French, respect the laws, and work. Today, a certain immigration is coming to deal shit, assault women, and disrespect France: we must send these people back to their home countries.] |
| **2. Identity attack:** Les Européens sont tellement faibles que tout le monde leur fait du chantage aux migrants. Tant qu'un pays comme la France ne dira pas : « les migrants, terminé », tout le monde le fera. [English: "Europeans are so weak that everyone uses migrant blackmail against them. As long as a country like France doesn't say, 'Migrants, it's over,' everyone will do it."] |
| **3. Insults:** Pendant 5 ans, Emmanuel Macron a soigneusement refusé d'« emmerder » les racailles, les gangs, les apprentis-djihadistes, les immigrés hors-la-loi, les antifas et les idéologues qui lavent les cerveaux de nos enfants. Lâche avec les forts, cruel avec les faibles. [English: For 5 years, Emmanuel Macron has carefully avoided 'bothering' troublemakers, gangs, apprentice jihadists, undocumented immigrants, antifas, and ideologues who brainwash our children. Weak with the strong, cruel with the weak.] |
| **4. Sexually explicit:** On marche sur la tête ! Un migrant afghan qui a violé un garçon de 12 ans à Saint-Brieuc est défendu sans vergogne par son avocat : le viol de garçons est une pratique courante en Afghanistan, l'esclavage sexuel y est répandu ! Nos dirigeants font la politique migratoire du chaos. [English: It's absurd! An Afghan migrant who raped a 12-year-old boy in Saint-Brieuc is shamelessly defended by his lawyer: the rape of boys is a common practice in Afghanistan, and sexual slavery is widespread! Our leaders are pursuing a chaotic immigration policy.] |
| **5. Threat:** A Montpellier, une trentaine de migrants albanais s'attaquent à des lycéens et poignardent un père de famille : STOP à la submersion migratoire et à l'impunité des clandestins ! Je réclame l'expulsion immédiate de ces voyous. [English: In Montpellier, some thirty Albanian migrants attacked high school students and stabbed a father: STOP the flood of migrants and the impunity of illegal immigrants! I demand the immediate expulsion of these thugs.] |

Fig 7: Examples of Tweets with Toxicity from each category in the French Tweets

Within the category *Identity attack,* we find examples where the identity of migrants is attacked, or about European identity but concerning the danger of migration (Ex. 2, Fig. 7). The category *Insults* targets migrants (Ex. 3, Fig. 7) but also other groups that bother the authors, such as humanitarian organisations ("les idiots utiles" - worthless idiots), intellectuals ("des bobos de Sciences Po", where bobo is a portmanteau of bourgeois-bohems (with the meaning similar to "champagne socialists"). The category *Sexually explicit* often draws a correlation between migrants and sexual crimes in France. However, in some instances, it is used to describe obscene everyday practices in the countries of origin of the migrants (Ex. 4, Fig. 7). The category *Threat* is, as expected, related to terrorism concerns in the context of global security issues as well as to higher crime rates that impact security (Ex. 5, Fig. 7).

*4.4 Persuasion in Tweets*

In the UK corpus, the most frequent persuasion techniques are Ad Hominem (19.2% of all tweets), followed by its subcategory Smears (13.4%), and next are Justification and Loaded language (both 9.4%); similarly for French Ad Hominem is followed by Justification (both 15%), Loaded language (11.15%) and Appeal to authority (8.3%). Ad Hominem and Smears both belong to the larger group of techniques related to attacking reputation. Ad Hominem attacks can refer either to concrete persons (such as "la communication lassante de Guéant" [Guéant's boring communication]) or institutions ("un gouvernement irrésolu et inerte" [an irresolute and inert government]) in the FR tweets. The category of Loaded Language also presents some interesting examples of specific words and phrases with strong emotional implications (Ex. 1, Fig. 8).

| |
|---|
| **1. Loaded Language:** Immigration minister Damian Green: This is a completely selfish and irresponsible decision by the #PCS leadership |
| **2. Justification:** En matière d'immigration clandestine, nous devons prendre l'Australie pour modèle, avec le renvoi automatique des clandestins, et le traitement des demandes d'asile dans les pays de départ. ⛔ Il est temps de dire « NO WAY » à nos frontières ! [English: In terms of illegal immigration, we must take Australia as a model, with the automatic deportation of undocumented individuals and the processing of asylum requests in the countries of origin. ⛔ It's time to say "NO WAY" to our borders!] |
| **3. Appeal to Fear:** Arrêtons, nous, Européens, d'être masochistes en versant des milliards d'euros à la #Turquie du sultan Erdogan qui, en retour, nous menace d'un chaos migratoire. Entre humiliation et honneur, il est temps de choisir, ou l'Histoire se chargera de le faire à notre place ! [English: Let us, Europeans, stop being masochistic by pouring billions of euros into #Turkey under Sultan Erdogan, who, in return, threatens us with a migration chaos. Between humiliation and honor, it is time to choose, or history will take care of it for us!] |

Fig 8: Examples of Persuasion Techniques detected in Tweets across both sub-corpora (UK, French)

The technique of Justification and Appeal to Authority both consist of two parts: a statement (to propose/support or not to propose/support something, like "Il est temps de dire « NO WAY » à nos frontières !") and an explanation for it (like "avec le renvoi automatique des clandestins, et le traitement des demandes d'asile dans les pays de départ") from Ex. 2, Fig. 8. Within the Justification group of techniques, we can find the technique Appeal to fear (Ex. 3, Fig. 8). We observe that not all the examples are correctly classified, however, the persuasion-techniques classifier is still a valuable tool for extracting examples for further qualitative analysis.

## 5. Conclusion and future work

Our findings highlight the effectiveness of NLP techniques in uncovering interpretable topic names and patterns within far-right discourse on Twitter. The correlation between identified topics and real-world events demonstrates the potential for these methods to offer valuable insights into the dynamics of right-wing ideologies in the digital space in a semi-automatic manner. Moreover, the toxicity and persuasion detection models shed light on the behavioural variations of nationalist groups across different countries and communities, providing a nuanced understanding of their online presence. In future, we plan to extend our analysis to left-wing perspectives (using Sangwan's (2023) comparable corpus) and discern differences in linguistic strategies between opposing ideological groups. Next, we would also apply the same methods to other languages and corpora covering a larger period. In addition, the three NLP methods used could be better integrated, e.g. enabling investigation of which topics and periods the toxicity is more present.

## 6. Acknowledgements

This work was supported by the Slovenian Research Agency grants via the core research programmes Knowledge Technologies (P2-0103) and Theoretical and Applied Linguistic Research: Contrastive, Synchronic, and Diachronic aspects (P6-0218), and the projects Computer-assisted multilingual news discourse analysis with contextual embeddings (J6-2581), Embeddings-based techniques for Media Monitoring Applications (L2-50070) and Hate Speech in Contemporary Conceptualizations of Nationalism, Racism, Gender and Migration (J5-3102).